\documentclass[letterpaper,journal]{IEEEtran}
\usepackage{amsmath,amsfonts}
\usepackage{algorithmic}
\usepackage{algorithm}
\usepackage{array}
\usepackage[caption=false,font=footnotesize,labelfont=sf,textfont=sf]{subfig}
\usepackage{textcomp}
\usepackage{stfloats}
\usepackage{url}
\usepackage{verbatim}
\usepackage{graphicx}
\usepackage{cite}
\usepackage{xcolor}
\usepackage{multirow}
\usepackage{enumitem}

\usepackage[hidelinks]{hyperref}

\def\etal{\textit{et al.}}
\def\ie{\textit{i.e.}}
\def\eg{\textit{e.g.}}

\newcommand{\argmax}{\operatorname*{arg\,max}}

\begin{document}

\title{Unfolding 3D Gaussian Splatting via\\ Iterative Gaussian Synopsis}

\author{Yuqin Lu, Yihua Dai, Yang Zhou, Guiqing Li, and Shengfeng He,~\IEEEmembership{Senior Member,~IEEE}
\thanks{This work was supported in part by the Guangdong Natural Science Funds for Distinguished Young Scholars (Grant 2023B1515020097), the Singapore Ministry of Education Academic Research Fund Tier 2 (Award No. MOE-T2EP20125-0016), the Singapore Ministry of Education Academic Research Fund Tier 1 (Proposal ID: 24-SIS-SMU-015), the Lee Kong Chian Fellowships, the National Natural Science Foundation of China (Grant 62472180), and the Natural Science Foundation of Guangdong (Grant 2024A1515011995). \textit{(Yuqin Lu and Yihua Dai contributed equally to this work.)} \textit{(Corresponding authors: Yang Zhou; Shengfeng He.)}}
\thanks{
     Yuqin Lu, Yihua Dai, and Yang Zhou are with the School of Computer Science and Engineering, South China University of Technology, Guangzhou, Guangdong 510006, China; they are also with the School of Computing and Information Systems, Singapore Management University, Singapore 188065 (e-mail: \href{mailto:matrixgle19@gmail.com}{matrixgle19@gmail.com}).
    }
    \thanks{
    Guiqing Li is with the School of Computer Science and Engineering, South China University of Technology, Guangzhou, Guangdong 510006, China.
    }
    \thanks{
    Shengfeng He is with the School of Computing and Information Systems, Singapore Management University, Singapore 188065 (e-mail: \href{mailto:shengfenghe@smu.edu.sg}{shengfenghe@smu.edu.sg}).
    }
}

\markboth{IEEE Transactions on Visualization and Computer Graphics}{Lu \MakeLowercase{\textit{et al.}}: Unfolding 3D Gaussian Splatting via Iterative Gaussian Synopsis}

\maketitle

\begin{abstract}
3D Gaussian Splatting (3DGS) has become a state-of-the-art framework for real-time, high-fidelity novel view synthesis. However, its substantial storage requirements and inherently unstructured representation pose challenges for deployment in streaming and resource-constrained environments. Existing Level-of-Detail (LOD) strategies, particularly those based on bottom-up construction, often introduce redundancy or lead to fidelity degradation.
To overcome these limitations, we propose \textit{Iterative Gaussian Synopsis}, a novel framework for compact and progressive rendering through a top-down ``unfolding'' scheme. Our approach begins with a full-resolution 3DGS model and iteratively derives coarser LODs using an adaptive, learnable mask-based pruning mechanism. This process constructs a multi-level hierarchy that preserves visual quality while improving efficiency.
We integrate hierarchical spatial grids, which capture the global scene structure, with a shared \textit{Anchor Codebook} that models localized details. This combination produces a compact yet expressive feature representation, designed to minimize redundancy and support efficient, level-specific adaptation. The unfolding mechanism promotes inter-layer reusability and requires only minimal data overhead for progressive refinement.
Experiments show that our method maintains high rendering quality across all LODs while achieving substantial storage reduction. These results demonstrate the practicality and scalability of our approach for real-time 3DGS rendering in bandwidth- and memory-constrained scenarios.
Project page: \url{https://cs-luyq.github.io/UnfoldingGS/}.
\end{abstract}

\begin{IEEEkeywords}
Novel view synthesis, 3D reconstruction, level-of-detail, progressive rendering.
\end{IEEEkeywords}

\begin{figure*}
    \centering
    \includegraphics[width=0.99\linewidth]{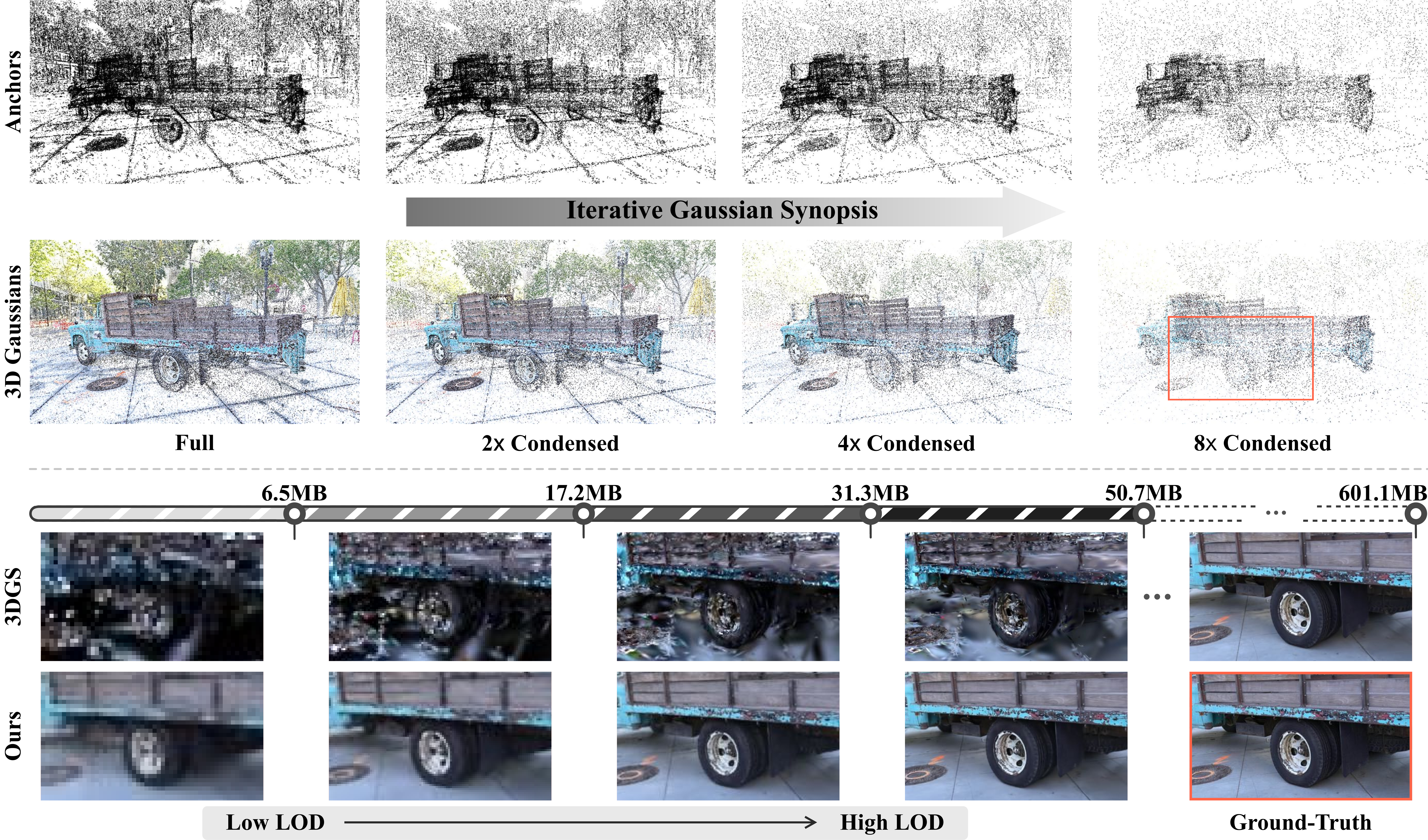}
    \caption{\textbf{Iterative Gaussian Synopsis} enables hierarchical ``unfolding'' of 3D Gaussian Splatting. The top rows illustrate the anchor structure alongside the derived 3D Gaussians at both full resolution and progressively condensed levels. The bottom rows compare the Level-of-Detail progression from a compact synopsis (starting at 6.5MB) to the highest fidelity (50.7MB) for our method, against a standard 3DGS baseline with a total size of 601.1MB. Our approach achieves clearer intermediate representations and significantly improved efficiency, delivering comparable visual quality at a fraction of the storage cost.}
    \label{fig:teaser}
\end{figure*}

\section{Introduction}
Recent advances in novel view synthesis and 3D scene representation have positioned 3D Gaussian Splatting (3DGS)~\cite{kerbl3Dgaussians} as a leading technique. By explicitly modeling scenes using millions of 3D Gaussian primitives and rendering them through a differentiable, tile-based rasterizer, 3DGS achieves real-time performance alongside state-of-the-art photorealistic quality. Unlike volumetric approaches such as NeRF~\cite{mildenhall2021nerf}, which require costly ray marching and sampling, 3DGS offers a highly efficient rendering pipeline. This combination of fidelity and efficiency has driven its adoption in diverse applications, including virtual and augmented reality, immersive telepresence, and large-scale scene visualization.

Despite its strengths, the practical deployment of 3DGS remains hindered by its substantial storage requirements. High-quality reconstructions typically involve millions of Gaussian primitives, resulting in model sizes that challenge real-time transmission, storage, and on-demand access, particularly in streaming or resource-constrained settings. To mitigate these issues, prior work has explored two main directions. One line of research focuses on compressing the Gaussian representation itself, reducing redundancy through parameter optimization or quantization~\cite{fan2023lightgaussian, chen2024hac, wang2024contextgs, ali2024trimming, lee2024compact, niedermayr2024compressed, navaneet2024compgs, scaffoldgs, liu2024hemgs, zhan2025cat}. Another stream of work aims to build Level-of-Detail (LOD) hierarchies~\cite{yan2024multi, kerbl2024hierarchical, ren2024octree, liu2025citygaussian}, allowing for scene representations that adapt to varying resource budgets or viewing conditions. However, both approaches face limitations in progressive rendering contexts, where a scene must be incrementally streamed and refined with maximal reuse of previously transmitted data.

Progressive rendering and streaming are increasingly critical for interactive 3D applications, where minimizing time-to-first-paint and enabling seamless detail refinement are essential. Several recent methods attempt to adapt 3DGS for such use cases. PRoGS~\cite{zoomers2024progs} proposes a heuristic importance-based ordering of Gaussians, prioritizing those with higher estimated contributions to early view synthesis. Octree-GS~\cite{ren2024octree} imposes a spatial hierarchy on the Gaussian primitives, enabling level-of-detail selection by spatial granularity. LapisGS~\cite{Shi2024LapisGS} introduces a layered training scheme where an initial coarse model, trained on low-resolution images, is iteratively refined with additional layers derived from high-resolution data. While these methods represent important steps toward progressive 3DGS rendering, they often rely on fixed ordering heuristics or bottom-up constructions that can limit their adaptability and efficiency. In particular, they may suffer from restricted inter-layer reuse, suboptimal integration of high-frequency details, and increased training or inference complexity due to cumulative optimization dependencies. As a result, existing solutions fall short of providing a unified and scalable framework for high-quality, bandwidth-aware progressive rendering across diverse operating conditions.

\begin{figure}
    \centering
    \includegraphics[width=\linewidth]{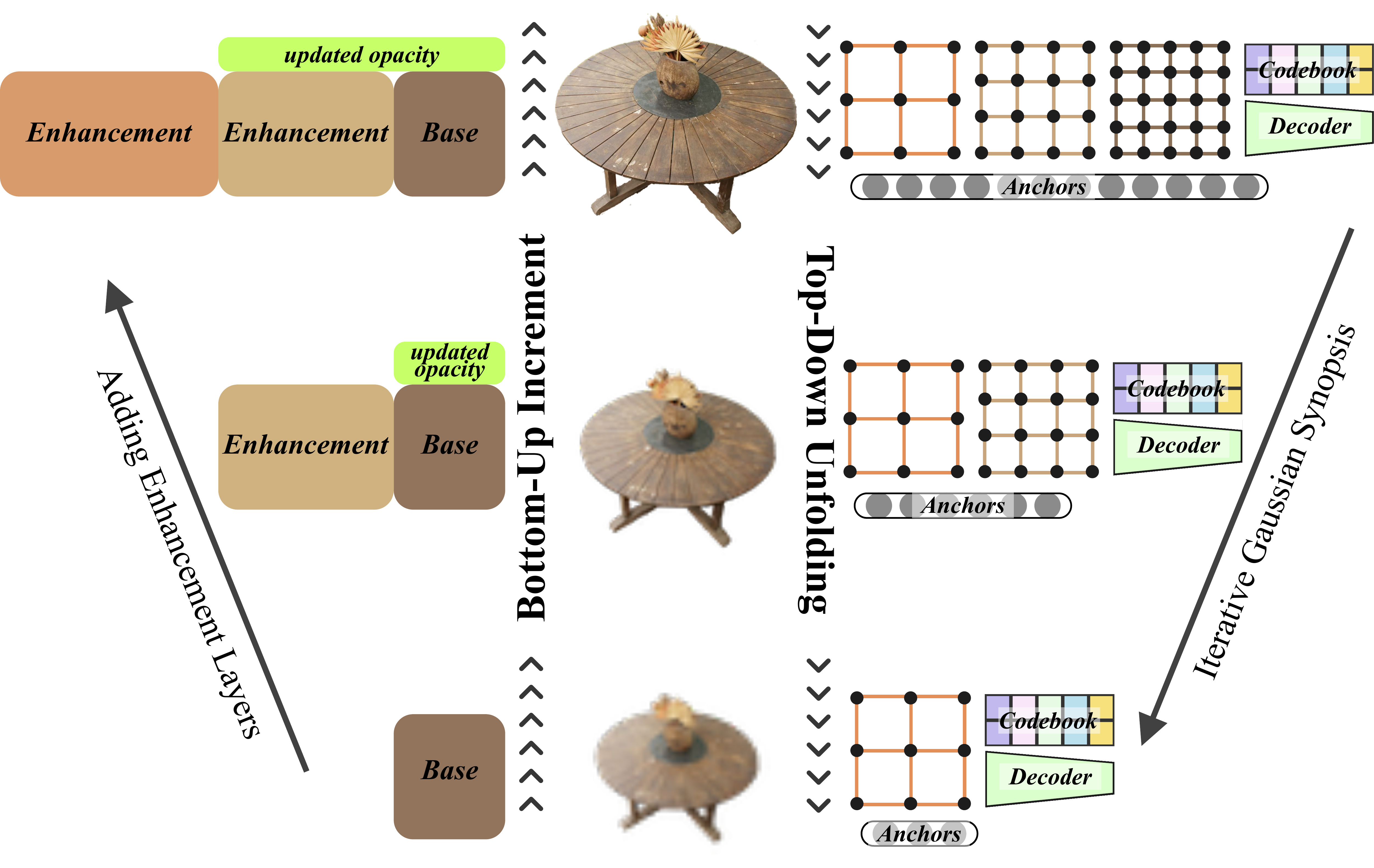}
    \caption{\textbf{Comparison of LOD Construction Strategies.} \textit{Left (Bottom-Up Increment):} A conventional bottom-up approach that builds from a coarse base layer by sequentially adding enhancement layers, typically requiring updates to the opacity or parameters of previous layers. \textit{Right (Top-Down Unfolding):} Our proposed top-down unfolding strategy, where lower levels of detail are systematically derived from a full-fidelity model, ensuring coherent simplification and improved inter-layer consistency.}
    \label{fig:difference}
\end{figure}

To address these limitations and provide a more robust solution for adaptive, high-fidelity rendering, we propose \textit{Iterative Gaussian Synopsis}, a novel top-down framework for constructing compact, multi-level LOD representations for 3DGS (see \figurename~\ref{fig:teaser}).
Our method conceptualizes LOD generation as a hierarchical "unfolding" process: beginning from a dense, full-resolution scene model and iteratively unpacking it to reveal progressively coarser, yet semantically coherent, representations embedded within.
This perspective stands in contrast to conventional bottom-up strategies that accumulate detail from sparse representations. Such approaches are often vulnerable to error propagation, representational drift, and optimization entanglement, especially when intermediate levels lack sufficient global context. By contrast, our method initiates from a globally optimized, high-fidelity model (illustrated in \figurename~\ref{fig:difference}), enabling subsequent LODs to be derived as structurally consistent simplifications that preserve key scene attributes.

\textit{Iterative Gaussian Synopsis} is anchored on three core design principles. First, we employ a hierarchical encoding architecture that enables both high compression and efficient unfolding. This is realized through multi-resolution hash grids that encode global scene structure, alongside a learnable \textit{Anchor Codebook} that models fine-grained local detail. Second, we construct the LOD hierarchy using a top-down, adaptive simplification process. Starting from the full 3DGS model, we generate lower LODs using a learnable, mask-guided pruning strategy that adaptively filters redundant primitives while preserving salient features. Each condensed state is thus a structurally faithful approximation of its parent. Third, to enhance both per-level fidelity and the overall compactness of the representation, we introduce two lightweight modules: \textit{Level-aware Decoding} (LAD), which allows flexible and resolution-sensitive feature decoding, and \textit{Coherent Basis Modulation} (CBM), which facilitates effective reuse of shared components across levels with minimal overhead.

Our framework enables rendering at various fidelity levels with minimal redundancy and maximal reuse of shared structure. This makes it particularly well-suited for progressive streaming, bandwidth-constrained scenarios, and dynamic level adaptation in interactive systems. Extensive experiments demonstrate that the proposed method achieves high-quality rendering across multiple levels of detail, while significantly reducing storage costs and enabling scalable, real-time 3DGS rendering.

\section{Related Work}
\subsection{Neural Rendering}
The advent of Neural Radiance Fields (NeRFs)~\cite{mildenhall2021nerf} has marked a major milestone in 3D neural rendering by modeling scenes as continuous volumetric functions, significantly advancing novel view synthesis.
While NeRF achieves remarkable rendering fidelity, its reliance on dense volumetric sampling and computationally intensive MLP evaluations results in slow training and inference. To mitigate these inefficiencies, subsequent works have explored explicit and structured scene representations~\cite{fridovich2022plenoxels, yu2021plenoctrees, muller2022instant, hedman2021snerg, chen2022tensorf, Reiser2021kilonerf, schwarz2022voxgraf, peng2021shape, xu2022point}, striking a balance between quality and efficiency. Recently, 3D Gaussian Splatting (3DGS)~\cite{kerbl3Dgaussians} has emerged as a compelling alternative. Unlike volumetric methods, 3DGS adopts a rasterization-based pipeline that optimizes a set of anisotropic 3D Gaussians to directly model the scene. This approach offers both superior visual quality and faster rendering performance, positioning 3DGS as a popular solution for real-time and high-fidelity neural rendering tasks.

\subsection{Compression of 3DGS}
3D Gaussian Splatting~\cite{kerbl3Dgaussians} has significantly advanced real-time scene rendering by employing explicit 3D Gaussian primitives. However, representing complex scenes with millions of primitives results in substantial storage demands and considerable bandwidth overhead.
Addressing this limitation, a significant body of research focuses on developing compression techniques for 3DGS to reduce model size and improve transmission efficiency. These approaches primarily aim to reduce the number of primitives and compress their parameters with quantization~\cite{niedermayr2024compressed, navaneet2024compgs, lee2024compact, xie2024sizegs}, pruning~\cite{ali2024trimming, fan2023lightgaussian, fang2024mini, papantonakis2024reducing, NEURIPS2024_lp3dgs, girish2023eagles}, spatial redundancy reduction~\cite{scaffoldgs, shin2025locality, zhang2024gaussian, chen2024hac, wang2024contextgs}, and entropy modeling~\cite{chen2024hac, zhan2025cat, wang2024contextgs, girish2023eagles, liu2024hemgs, wang2024end}.
For instance, Niedermayr \etal ~\cite{niedermayr2024compressed} achieved effective 3DGS compression through a combination of sensitivity-aware vector clustering, quantization-aware fine-tuning, and entropy encoding. Scaffold-GS~\cite{scaffoldgs} employed spatial correlation to cluster neighboring Gaussians into anchors and encode their attributes using MLPs. Building upon this, methods like HAC and ContextGS~\cite{chen2024hac, wang2024contextgs, zhan2025cat, liu2024hemgs} leverage entropy coding or context modeling to further exploit spatial consistencies for more efficient representations. While effective in reducing the absolute model size, these static compression methods do not inherently support progressive transmission or rendering. The heavy reliance on context-adaptive entropy decoding introduces sequential computational overhead that is hostile to real-time streaming, and the requirement to download the entire compressed payload before rendering violates the ``time-to-first-paint'' mandate. Furthermore, retraining is often required to target different quality levels, making them inefficient for dynamic streaming scenarios. In contrast, our Iterative Gaussian Synopsis framework is explicitly designed to solve the progressive LOD generation problem with highly parallelizable MLP decoding. We view advanced static compression techniques like HAC~\cite{chen2024hac} and HAC++~\cite{chen2025hac++} as highly complementary, as their entropy coding principles could potentially be integrated to further compress the delta payloads between our nested LODs in future end-to-end streaming pipelines.

\subsection{Level-of-Detail 3DGS}
Level-of-Detail rendering ~\cite{hoppe1997progressivemeshes,chen2023progressivemeshes, nglod, xiangli2022bungeenerf} is essential for efficiently managing complex scenes by adapting the rendered detail based on viewpoint or resource availability.
Adapting LOD concepts to 3DGS is crucial for scalability and progressive rendering, especially in large-scale or bandwidth-constrained environments. Several approaches have introduced computer graphics concepts like mipmapping and hierarchical structures to 3DGS~\cite{yan2024multi, kerbl2024hierarchical}, aiming to improve rendering efficiency and mitigate aliasing. PRoGS~\cite{zoomers2024progs} introduced a simple approach for progressive rendering by ordering Gaussians based on their contribution to the final scene.
Octree-GS~\cite{ren2024octree} organizes scenes with an octree structure, allowing dynamic selection of LOD levels based on viewing distances. However, these methods define LODs based on spatial partitioning or independent scale-specific models
without utilizing correlation between levels, which can involve redundant storage and hinder a smoothly progressive representation.

More recently, LapisGS~\cite{Shi2024LapisGS} introduced a layered structure for cumulative representation, which builds a base layer from low-resolution images and adds enhancement layers, optimizing opacity of prior layers to mediate contributions.
This cascaded training scheme that LapisGS~\cite{Shi2024LapisGS} adopts can compromise fidelity and introduce optimization conflicts between layers, resulting in accumulated errors from earlier stages. In contrast, our Iterative Gaussian Synopsis is poised to overcome the inherent error accumulation and optimization limitations of bottom-up approaches. This offers the potential for a more efficient and consistent scene representation while delivering high-quality multi-level LOD rendering.

\begin{figure*}
    \centering
    \includegraphics[width=\linewidth]{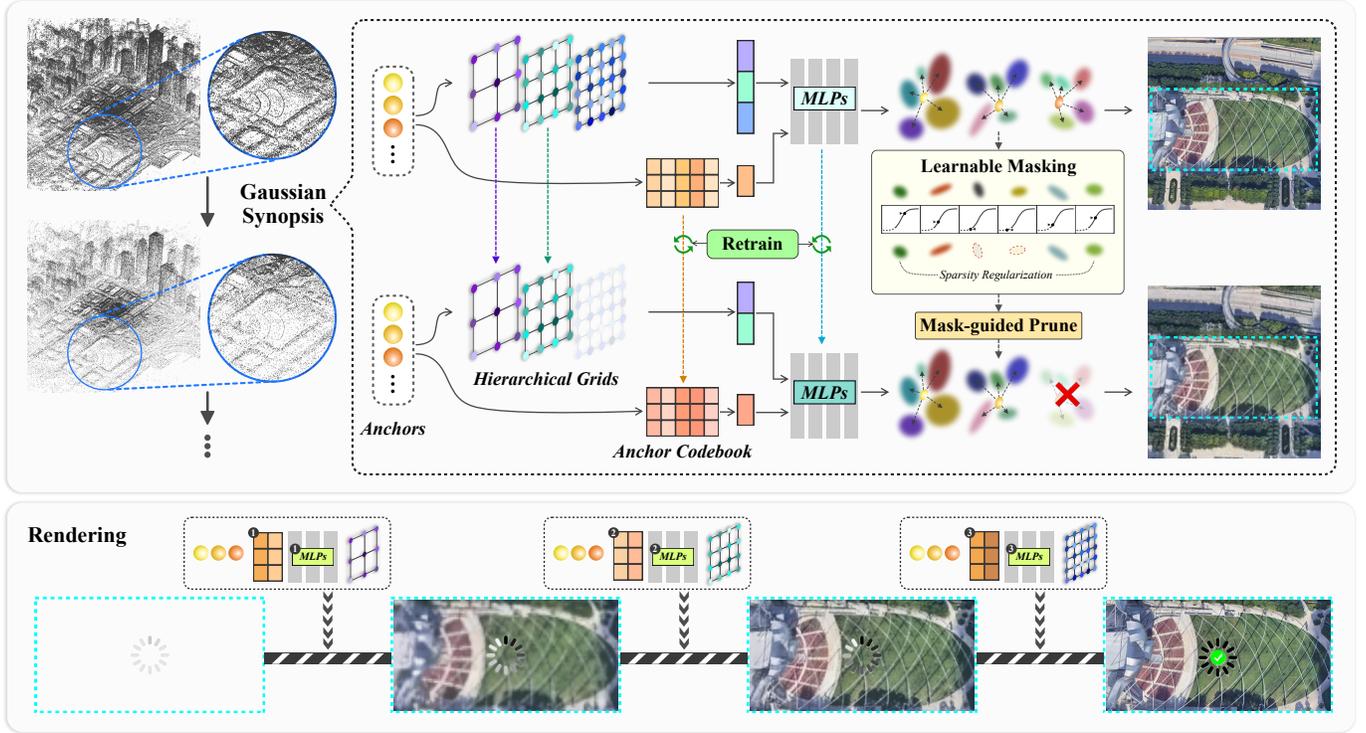}
    \caption{\textbf{Overview of the Iterative Gaussian Synopsis Framework.}
\textit{Top:} The full-resolution 3DGS model is progressively ``unfolded'' into coarser LODs using a learnable, mask-guided pruning strategy. \textit{Bottom:} The resulting hierarchical LODs support efficient progressive rendering, where visual quality improves smoothly as more data is streamed, transitioning seamlessly from a coarse structural outline to a high-fidelity reconstruction.}\label{fig:synopsis_arch}
\end{figure*}

\section{Method}
\subsection{Preliminaries}
{\bf 3D Gaussian Splatting}~\cite{kerbl3Dgaussians} represents a 3D scene using a collection of anisotropic 3D Gaussians, enabling real-time, high-quality rendering through efficient tile-based rasterization. Each 3D Gaussian is defined by its spatial position and a covariance matrix that describes its anisotropic extent in 3D space:
\begin{equation*}
G(x) = \exp\left( -\frac{1}{2}(x - \mu)^\top \Sigma^{-1}(x - \mu) \right),
\end{equation*}
where $\mu$ denotes the Gaussian center, and $\Sigma$ is the 3D covariance matrix. As $\Sigma$ is positive semi-definite, it admits a decomposition $\Sigma = RS S^\top R^\top$ with a rotation matrix $R$ and a diagonal scaling matrix $S$.
In addition to its geometric parameters, each Gaussian is associated with an opacity value $\alpha$ and a set of Spherical Harmonics (SH) coefficients that represent its view-dependent color $c$.

To render an image from a given viewpoint, each 3D Gaussian is first projected onto the 2D image plane using an affine approximation of the camera projection matrix, resulting in a 2D elliptical Gaussian with transformed mean and covariance. A tile-based rasterizer then sorts these projected Gaussians in front-to-back depth order and applies alpha compositing to accumulate color contributions. Specifically, the final color at a pixel $x'$ is computed as follows:
\begin{equation*}
C(x') = \sum_{i \in \mathcal{N}(x')} c_i \cdot \sigma_i(x') \cdot \prod_{j=1}^{i-1} (1 - \sigma_j(x')),
\end{equation*}
where $\sigma_i(x') = \alpha_i G_i'(x')$ denotes the weighted 2D density, $c_i$ is the view-dependent color obtained via Spherical Harmonics, and $\mathcal{N}(x')$ is the set of Gaussians that contribute to pixel $x'$, sorted by depth. This formulation allows 3DGS to efficiently approximate volumetric rendering with high fidelity and real-time performance.

{\bf Scaffold-GS}~\cite{scaffoldgs} introduces a more compact representation for 3D Gaussians by predicting them from a sparse set of anchor points, thereby significantly reducing storage overhead. Instead of storing the properties of every 3D Gaussian individually, Scaffold-GS voxelizes the 3D scene and initializes each voxel with an anchor point that contains a feature $\mathbf{f}_a \in \mathbb{R}^{32}$, a position $\mathbf{x}_a \in \mathbb{R}^3$, a base scaling factor $\mathbf{l}_a \in \mathbb{R}^3$, and $k$ learnable offsets $\mathbf{O}_a \in \mathbb{R}^{k \times 3}$.

Given a camera position $\mathbf{x}_c$, each anchor point uses the associated features $\mathbf{f}_a$, the relative viewing distance $\sigma_c$ and direction $\vec{d}_c$ to predict the properties of $k$ neural Gaussians within its voxel via neural prediction:
\begin{equation*}
    \{\alpha^i, \mathbf{c}^i, \mathbf{s}^i, \mathbf{r}^i\}_{i=0}^k= \mathcal{F}(\mathbf{f}_a, \sigma_c, \vec{d}_c),
\end{equation*}
with
\begin{equation*}
    \sigma_c = \|\mathbf{x} - \mathbf{x}_c\|_2 ,\quad \vec{d}_c = \frac{\mathbf{x} - \mathbf{x}_c}{\|\mathbf{x} - \mathbf{x}_c\|_2}.
\end{equation*}
Here, $\mathcal{F}$ contains four individual MLPs, $F_\alpha$, $F_{\mathbf{c}}$, $F_\mathbf{s}$ and $F_\mathbf{r}$, for predicting opacity $\{\alpha^i\}$, color $\{\mathbf{c}^i\}$, scale $\{\mathbf{s}^i\}$ and rotation $\{\mathbf{r}^i\}$ respectively for the spawned Gaussians. The 3D positions of the Gaussians are then computed as $\{\boldsymbol{\mu}^i\} = \mathbf{x}_a + \{\mathbf{O}_a^i\} \cdot \mathbf{l}_a$, making them spatially distributed around their anchor. Once these properties are decoded, the rendering process follows the same pipeline as 3DGS. By encoding only the anchor features and predicting full Gaussian attributes at inference time, Scaffold-GS dramatically reduces memory usage while preserving rendering quality.

\begin{figure}
    \centering
    \includegraphics[width=\linewidth]{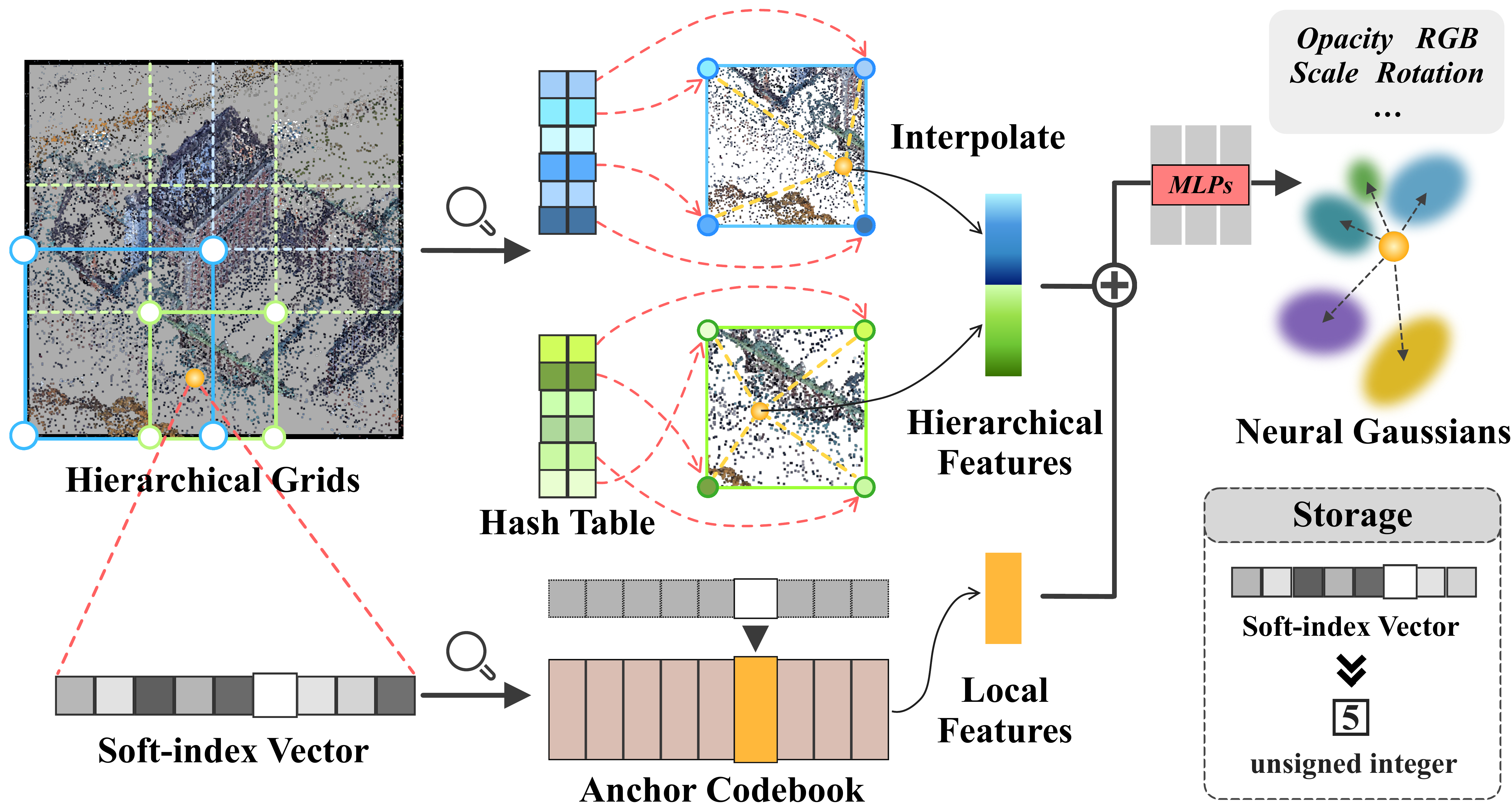}
    \caption{\textbf{Our hierarchical architecture.} The hierarchical grids capture multi-scale contextual information while the learnable codebook models localized details, formimg the foundation for unfolding into LODs.}
    \label{fig:hierarchy_arch}
\end{figure}

\subsection{Foundational Architecture}\label{sec:arch}

\figurename~\ref{fig:synopsis_arch} shows an overview of our method. As illustrated, the capacity of our Iterative Gaussian Synopsis framework to effectively ``unfold'' into meaningful LODs hinges on a robust and hierarchically structured 3DGS representation. Our method is built upon an anchor-based structure, where anchors, positioned at $x_{a}\in\mathbb{R}^{3}$, serve as central points for parameterizing local groups of 3D Gaussians. These anchors are typically initialized from SfM~\cite{schonberger2016structure} point clouds after a voxelization step, as in Scaffold-GS~\cite{scaffoldgs}.

The construction of our hierarchical architecture is depicted in~\figurename~\ref{fig:hierarchy_arch}. To embed multi-scale global context into each anchor,
we utilize features derived from $L_g$ levels of multi-resolution hash grids~\cite{muller2022instant}. For an anchor at position $x_{a}$, the hierarchical feature is obtained by concatenating the interpolated features across all grid levels, denoted as $f_{\text{hier}}(x_a) = [ \text{Interpolate}(\mathcal{G}_g, x_a) ]_{g=1}^{L_g}$, where $\mathcal{G}_g$ represents the $g$-th grid level and $\text{Interpolate}(\cdot)$ denotes multi-linear interpolation at $x_a$.
This component equips the anchor representation with the multi-scale awareness necessary for our subsequent top-down unfolding scheme.

To complement this global context with fine-grained local variations, we introduce a global, learnable Anchor Codebook $C\in \mathbb{R}^{N_c\times D_c}$. This design choice is favored as it provides a mechanism to introduce rich, localized features that can be made level-specific without incurring substantial redundancy across levels.
Specifically, each anchor $j$ uses a learnable soft-index vector $v_{j} \in \mathbb{R}^{N_c}$ and a straight-through estimator to retrieve a local feature $f_{\text{loc},j} \in \mathbb{R}^{D_c}$  during training.
We adopt the straight-through estimator~\cite{bengio2013estimating} to circumvent the non-differentiable indexing problem,  formulated as:
$$w_j = \text{Softmax}(v_j),$$
$$f_{\text{loc},j} = [\text{Onehot}(\text{argmax}(w_j^T)) + w_j^T -\text{sg}(w_j^T)] \cdot C$$
where $\text{sg}(\cdot)$ denotes the stop-gradient operation. This allows each anchor to select or synthesize a specific local feature that best complements its broader hierarchical context. For efficient storage and deterministic fast retrieval during inference, the soft-index $v_j$ is converted to a discrete integer index $\text{idx}_j = \text{argmax}(v_j)$, and the local feature is directly fetched as $f_{\text{loc},j} = C[\text{idx}_j]$.

The hierarchical and local features are then concatenated and processed by a fusion MLP:
$$F(x_a) = \text{MLP}_{\text{fuse}}(f_{\text{hier}}(x_a) \oplus f_{\text{loc}})$$
This fused feature is passed to attribute-prediction MLPs ($\text{MLP}_\text{attr}$) to decode the final opacity, scale, rotation, and color for the generated Gaussians. The initial full-fidelity model is trained end-to-end to minimize the loss function $\mathcal{L}_{\text{full}}$, composed of an L1 photometric loss, a D-SSIM term, and a volume regularization term $\mathcal{L}_{\text{vol}}$ on the Gaussian scales:
$$\mathcal{L}_{\text{full}} = (1-\lambda_{\text{SSIM}})\mathcal{L}_1 + \lambda_{\text{SSIM}}\mathcal{L}_{\text{D-SSIM}} + \lambda_{\text{vol}}\mathcal{L}_{\text{vol}}.$$
The volume term discourages degenerate over-large Gaussians by penalizing their volume, $\mathcal{L}_{\text{vol}} = \mathbb{E}_i[\prod_{d} s^{i}_{d}]$, the product of the three entries of the predicted scale $\mathbf{s}^i$.
This full-fidelity model serves as the highest LOD and the foundation for our subsequent top-down unfolding.

\subsection{Iterative Gaussian Synopsis}
Our \textit{Iterative Gaussian Synopsis} framework employs a distinct top-down strategy to construct a hierarchy of LOD representations. This iterative process begins with the comprehensive full-fidelity model as the finest LOD and progressively derives coarser levels through an adaptive, learnable pruning mechanism. This approach is designed to ensure that each coarser LOD is a coherent and salient simplification of its preceding, more detailed level, while also adapting the underlying feature extraction machinery to the target level of detail.

The full-fidelity representation, established using the hierarchical anchor features, constitutes the highest level of detail, denoted as LOD $(L_{max}-1)$. Subsequent, coarser levels, LOD $(L_{max}-2)$, $(L_{max}-3)$, ..., LOD 0, are then derived sequentially: LOD $\ell$ is condensed by applying our Gaussian synopsis strategy to generate LOD $(\ell-1)$.

\subsubsection*{\bf Hierarchical Grids Downsampling}
The first step in our unfolding process involves adapting the hierarchical grids that provide global context. As we iteratively ``unfold'' to coarser LODs that require less fine-grained detail, the fine-grained contextual details provided by higher-resolution grid levels become less critical and can be computationally expensive. Therefore, we implement a progressive downsampling of the hierarchical grids, where higher-resolution grid levels are iteratively deactivated for coarser LODs.

Specifically, if the full-fidelity model utilizes a set of $L_g$ grid levels $\{\mathcal{G}_g\}_{g=1}^{L_g}$ for feature extraction, then with $L_{max}$ LODs, unfolding from $\ell$ to $\ell-1$ discards the finest $L_g/L_{max}$ grid levels, so each coarser LOD reuses only the coarser part of the grids allocated at full fidelity rather than allocating new ones. This ensures that the feature extraction process for each LOD is commensurate with its detail requirements, reducing computational overhead and focusing the feature representation on the appropriate scale for that specific level.

\subsubsection*{\bf Learnable Anchor Pruning}
The core mechanism for condensing the scene representation at each unfolding step is an adaptive, learnable mask-based pruning strategy applied to the anchors. This strategy determines which anchors are retained or pruned when transitioning from a finer LOD to the next coarser one, effectively controlling the density and complexity of the representation at each level. Our approach is inspired by learnable masking techniques for individual Gaussians, but adapted to our anchor-based framework to encourage structurally coherent pruning.

For each anchor $j$ that generates a set of $K$ Neural Gaussians, we introduce a corresponding learnable mask parameter vector $\mathbf{m}_j\in \mathbb{R}^K$. Each element $m_{j,i}$ corresponds to the $i$-th Gaussian spawned by the anchor. Following ~\cite{lee2024compact}, we obtain a binary mask value $M_{j,i} \in \{0,1\}$ for each potential Gaussian using the sigmoid function $\sigma(\cdot)$ and a straight-through estimator (STE)~\cite{bengio2013estimating} to enable gradient propagation during training:
$$M_{j,i} = \text{sg}(\mathbf{1}[\sigma(m_{j,i}) > \epsilon] - \sigma(m_{j,i})) + \sigma(m_{j,i})$$
where $\text{sg}(\cdot)$ is the stop-gradient operator, $\mathbf{1}[\cdot]$ is the indicator function, and $\epsilon$ is a constant threshold. Considering the contribution to a condensed state and the detail granularity, this binary mask is specifically applied to the opacity $\alpha_{j,i}$ and scale vector $\mathbf{s}_{j,i}$ of the generated Gaussians:
$$\hat{\alpha}_{j,i} = M_{j,i} \cdot \alpha_{j,i} \quad \text{and} \quad \hat{\mathbf{s}}_{j,i} = M_{j,i} \cdot \mathbf{s}_{j,i}.$$
An anchor $j$ is considered pruned or inactive for a given LOD if all of its associated masks are zero (\ie, $\sum_{i=1}^K M_{j,i}=0$).

\subsubsection*{\bf Sparsity Regularization for Pruning}
To guide the learnable pruning process and encourage the model to derive condensed representations, we apply a sparsity regularization term, $\mathcal{L}_\text{sparsity}$ to the learnable mask parameters $\mathbf{m}_j$. This regularization consists of two distinct components: a standard per-Gaussian masking loss and our group sparsity loss, which work in tandem to promote both fine-grained and structurally coherent pruning. The standard masking loss ~\cite{lee2024compact} encourages the deactivation of any single non-essential Gaussian, normalized over all $NK$ mask entries:
$$\mathcal{L}_{\text{mask}} = \tfrac{1}{NK}\textstyle\sum_{j}\sum_{i=1}^K \sigma(m_{j,i}).$$

While this loss promotes general sparsity, it does not explicitly encourage the removal of entire anchors, which is critical for achieving significant storage savings in our unfolding scheme. To achieve more structurally coherent pruning, we introduce a group sparsity loss, $\mathcal{L}_\text{group}$, which is specifically designed to encourage the entire group of Gaussians associated with a single anchor to be "switched off" together. We treat the vector of mask activations for each anchor $j$, $\sigma(\mathbf{m}_j)$, as a single group and penalize its L2 norm averaged over the $N$ anchors:
$$\mathcal{L}_{\text{group}} = \tfrac{1}{N}\textstyle\sum_{j} ||\sigma(\mathbf{m}_j)||_2.$$

This group-wise regularization is more effective at pruning entire anchors cohesively compared to an L1-style penalty, which could leave anchors with only a few sparsely active Gaussians, hindering the compactness of our condensed model. The total sparsity regularization is a weighted sum of these two components:
$$\mathcal{L}_{\text{sparsity}} = \lambda_{\text{mask}}\mathcal{L}_{\text{mask}} + \lambda_{\text{group}}\mathcal{L}_{\text{group}}$$
where $\lambda_{\text{group}}$ is a hyperparameter that balances the two regularization terms. Both terms are normalized so that the regularization strength is invariant to scene size. Each unfolding step optimizes $\mathcal{L}_{\text{LOD}} = \mathcal{L}_{\text{full}} + \mathcal{L}_{\text{sparsity}}$.

\subsection{Level-Specific Adaptation}
Once a condensed LOD is structurally defined through our top-down unfolding and pruning scheme, the representation of the anchors must be optimized to ensure high fidelity for that specific level.
Rather than redefining the entire feature extraction pipeline, we adapt the foundational architecture using two lightweight modules.

\subsubsection*{\bf Coherent Basis Modulation}
Instead of using a single, static codebook across all LODs, Coherent Basis Modulation produces a level-specific version of the codebook, $C^{(\ell)}$, for LOD $\ell$. This is achieved by fine-tuning or modulating the entries of the global, foundational codebook $C$ for each specific detail level. For a given anchor $j$, its local feature representation is adapted for LOD $\ell$ by retrieving the corresponding modulated entry using its fixed foundational index $\text{idx}_j$:
 $$f_{\text{loc},j}^{(\ell)} = C^{(\ell)}[\text{idx}_j].$$
Since the "Anchor Codebook" itself is highly compact, the cost of storing the fine-tuned adjustments or even a distinct version for each LOD represents a negligible storage increment, avoiding the massive redundancy of storing unique local features for every anchor at every level. Crucially, since the index is fixed, this module preserves the learned structural relationships between anchors while allowing their shared feature ``basis'' to be tailored to the current detail requirements. At full fidelity the codebook is randomly initialized. When unfolding, the table of the target level is copied from the trained foundational codebook and then fine-tuned, which preserves coherence of the shared basis.

\subsubsection*{\bf Level-aware Decoding}
To interpret the modulated features optimally for each level's objective, Level-aware Decoding introduces a level-specific set of fusion and attribute-prediction MLPs ($\text{MLP}_{\text{fuse}}^{(\ell)}$ and $\text{MLP}_{\text{attr}}^{(\ell)}$). These MLPs are fine-tuned specifically for LOD $\ell$ and applied to the downsampled hierarchical grids and modulated local features. By adapting a compact, function-based predictor rather than storing explicit Gaussian attributes for each level, LAD enables high fidelity and structural flexibility while maintaining strict model compactness. Algorithm~\ref{alg:unfold} collects these components into the complete procedure: full-fidelity training yields the highest LOD and the hardened indices, and each unfolding step then restricts the grids, seeds the level-specific codebook and MLPs from their foundational counterparts, and fine-tunes under $\mathcal{L}_{\text{LOD}}$ until the anchors condense to the next coarser level.

\begin{algorithm}[t]
\caption{Full-fidelity training and top-down unfolding}
\label{alg:unfold}
\footnotesize
\begin{algorithmic}[1]
\REQUIRE SfM points; images at the four target resolutions
\STATE Initialize anchors by voxelization; offsets, scales, MLPs, grids, codebook $C$, soft indices $\{v_j\}$
\FOR{$t = 1 \dots T_{\text{full}}$}
  \STATE Render a sampled view; minimize $\mathcal{L}_{\text{full}}$; update all parameters
\ENDFOR
\STATE $\mathrm{idx}_j \leftarrow \argmax(v_j)$; this model is LOD~L3
\FOR{$\ell = 3$ \textbf{downto} $1$}
  \STATE Restrict grids to LOD $\ell-1$; masks $\leftarrow 1$; $C^{(\ell-1)} \leftarrow C$; $\text{MLP}^{(\ell-1)} \leftarrow \text{MLP}$
  \FOR{$t = 1 \dots T_{\text{unfold}}$}
    \STATE $M_{j,i} \leftarrow \mathrm{sg}(\mathbf{1}[\sigma(m_{j,i})>\epsilon]-\sigma(m_{j,i})) + \sigma(m_{j,i})$; $\hat\alpha \leftarrow M\alpha$; $\hat s \leftarrow Ms$
    \STATE Render at the target resolution of LOD $\ell-1$; minimize $\mathcal{L}_{\text{LOD}}$ updating the masks and, per LAD/CBM, the level-specific MLPs and $C^{(\ell-1)}$; freeze positions, offsets, scales and grids
  \ENDFOR
  \STATE Delete anchors with $\sum_i M_{j,i}=0$; emit compressed delta
\ENDFOR
\RETURN $\{\text{L0},\dots,\text{L3}\}$
\end{algorithmic}
\end{algorithm}

\section{Experiments}
\subsection{Experimental Setups}\label{sec:setup}
This section details the experimental setup used to validate our proposed Iterative Gaussian Synopsis framework. We outline the datasets, baseline methods for comparison, evaluation metrics, and specific implementation details of our approach.

\subsubsection*{\bf Datasets}
We conduct a comprehensive evaluation of our method across several public datasets to demonstrate its effectiveness in diverse scenarios, including Mip-NeRF360~\cite{barron2021mip}, Tanks \& Temples~\cite{knapitsch2017tanks}, DeepBlending~\cite{hedman2018deep}, and BungeeNeRF~\cite{xiangli2022bungeenerf}. These datasets encompass complex indoor and outdoor scenes that challenge rendering quality and detail preservation. All methods are evaluated on an identical scene set per dataset, which for Mip-NeRF360 comprises all nine scenes.

\subsubsection*{\bf Baseline Methods}
We compare our method against several state-of-the-art approaches to benchmark its performance in terms of rendering quality, model compactness, and LOD capabilities. We provide comparisons with methods that support LOD rendering, including Octree-GS~\cite{ren2024octree} and LapisGS~\cite{Shi2024LapisGS}, to specially evaluate our LOD unfolding scheme. For the original 3DGS~\cite{kerbl3Dgaussians}, we re-order its Gaussians based on their contributions~\cite{zoomers2024progs} to enable progressive support and render the scene respectively with 1/8, 1/4, 1/2, and all Gaussian splats. For Octree-GS, we fix the number of octree levels to four for all scenes to match our four LODs, instead of using the default level count derived per scene. The anchors of depth $d$ are the centers of voxels of side $V/2^{d}$, where the base size $V$ follows from the scene bounding box. The original method selects the depths to render for each view from the viewing distance and reports no per-level results, so we define four operating points as depth cutoffs: L0 to L3 render only the anchors of depths $0$ to $d$ for $d=0,1,2,3$ respectively. L0 therefore keeps only the coarsest voxel population, while L3 retains all depths of our fixed four-level configuration and is not identical to the default model.

\subsubsection*{\bf Evaluation Metrics}
Our evaluation employs a set of standard metrics to assess visual fidelity, model compactness, and streaming efficiency. We measure rendering quality using Peak Signal-to-Noise Ratio (PSNR), Structural Similarity Index Measure (SSIM), and Learned Perceptual Image Patch Similarity (LPIPS). To evaluate the framework's suitability for streaming, we measure the total model size (in megabytes, MB) alongside the Compressed Delta Size, which quantifies the incremental network payload required to transition from one LOD to the next. Finally, to ensure practical deployment viability, we report the rendering speed in Frames Per Second (FPS).

\subsubsection*{\bf Implementation Details}
Our framework is implemented in PyTorch, building upon the official repository of Scaffold-GS~\cite{scaffoldgs}. We generate a hierarchy of 4 LODs (L0 to L3). For the hierarchical grids, we use a multi-resolution hash grid with 12 levels and a feature dimension of 4. The global Anchor Codebook size $N_c$ is set to 256 with a feature dimension of $D_c = 32$. The group sparsity loss weight $\lambda_\text{group}$ is set to 0.01. All models are trained for 30,000 iterations at full fidelity and iteratively unfold to coarser LODs with another 10,000 iterations on a single NVIDIA L40 GPU. For our framework, the training loss weights are set to $\lambda_\text{SSIM} = 0.2$ and $\lambda_\text{vol}= 0.001$ to optimally balance photometric and structural accuracy for our shared-codebook architecture. Except for the fixed Octree-GS level count and the progressive evaluation adaptations described above, we follow the official implementations and default hyperparameter configurations of all baseline methods (\eg, $\lambda_\text{SSIM} = 0.8$ for LapisGS~\cite{Shi2024LapisGS}).

The hierarchical grids are initialized uniformly in $[-10^{-4},10^{-4}]$ and codebook entries from $\mathcal{N}(0,1)$. LOD $\ell$ retains the first $3(\ell+1)$ of the 12 grid levels, i.e.\ 12, 9, 6 and 3 for L3 to L0. We set $\epsilon=0.01$ and $\lambda_\text{mask}=0.005$, with all masks initialized to one. During each unfolding step the anchor positions, offsets and base scales remain frozen, as do the hierarchical grids, which are trained once at full fidelity. The masks are updated in every step, LAD additionally updates the level-specific fusion and attribute-prediction MLPs, and CBM additionally updates the codebook table of the target level. The learning rates are 0.01 for the masks, 0.005 for the fusion MLP, 0.002, 0.008 and 0.004 for the opacity, color and covariance MLPs, and 0.005 for the codebook.

To accurately simulate the streaming payload, the compressed delta size between LODs is computed by isolating the newly activated parameters (including un-pruned anchors, novel grid levels, level-specific MLP weights and Codebook) and applying standard compression utilizing the DEFLATE algorithm. Following LapisGS~\cite{Shi2024LapisGS}, we calculate image quality metrics for lower LODs at their proportionally scaled target resolutions; all reported comparisons use held-out test views.

\begin{table*}[t]
    \centering
    \caption{Quantitative comparisons of our method against 3DGS~\cite{kerbl3Dgaussians}, Octree-GS (Oct-GS)~\cite{ren2024octree}, and LapisGS~\cite{Shi2024LapisGS} across different datasets and LOD levels. Best and second-best results are highlighted in \colorbox{red!40}{red} and \colorbox{yellow!40}{yellow}, respectively. Size is measured in MB. It should be noted that the reported  sizes for 3DGS and LapisGS represent their default, uncompressed explicit Gaussian representations. Following LapisGS, metrics at each LOD are computed at the proportionally scaled target resolution of that level for every method. Coarser levels therefore face easier targets and are comparable only within a level.}
    \label{tab:lod_comparisons}
    \footnotesize
    \renewcommand{\arraystretch}{1.4}
    \setlength\tabcolsep{1.4pt}
        \begin{tabular}{c|l|cccc|cccc|cccc|cccc}
        \hline
        \multirow{2}{*}{Level} & \multirow{2}{*}{Metric} & \multicolumn{4}{c|}{Mip-NeRF360~\cite{barron2021mip}} & \multicolumn{4}{c|}{BungeeNeRF~\cite{xiangli2022bungeenerf}} & \multicolumn{4}{c|}{DeepBlending~\cite{hedman2018deep}} & \multicolumn{4}{c}{Tanks\&Temples~\cite{knapitsch2017tanks}} \\
        \cline{3-18}
        & & 3DGS & Oct-GS & LapisGS & Ours & 3DGS & Oct-GS & LapisGS & Ours & 3DGS & Oct-GS & LapisGS & Ours & 3DGS & Oct-GS & LapisGS & Ours \\
        \hline
        \multirow{6}{*}{L0}
        & PSNR$\uparrow$ & 18.39 & 17.44 & \colorbox{red!40}{28.10} & \colorbox{yellow!40}{28.02} & 17.05 & 19.20 & \colorbox{yellow!40}{26.23} & \colorbox{red!40}{27.23} & 20.55 & \colorbox{yellow!40}{22.56} & 19.99 & \colorbox{red!40}{27.95} & 15.10 & 12.55 & \colorbox{yellow!40}{23.41} & \colorbox{red!40}{23.68} \\
        & SSIM$\uparrow$ & 0.465 & 0.447 & \colorbox{red!40}{0.869} & \colorbox{yellow!40}{0.860} & 0.473 & 0.594 & \colorbox{yellow!40}{0.886} & \colorbox{red!40}{0.912} & 0.614 & \colorbox{yellow!40}{0.720} & 0.569 & \colorbox{red!40}{0.861} & 0.492 & 0.342 & \colorbox{yellow!40}{0.864} & \colorbox{red!40}{0.865} \\
        & LPIPS$\downarrow$ & 0.380 & 0.446 & \colorbox{red!40}{0.117} & \colorbox{yellow!40}{0.134} & 0.384 & 0.293 & \colorbox{yellow!40}{0.108} & \colorbox{red!40}{0.106} & 0.324 & \colorbox{yellow!40}{0.282} & 0.374 & \colorbox{red!40}{0.142} & 0.347 & 0.512 & \colorbox{red!40}{0.119} & \colorbox{yellow!40}{0.123} \\
        & FPS$\uparrow$ & \colorbox{red!40}{189} & 180 & 151 & \colorbox{yellow!40}{183} & 157 & 132 & \colorbox{yellow!40}{158} & \colorbox{red!40}{159} & \colorbox{red!40}{191} & 182 & 125 & \colorbox{yellow!40}{188} & \colorbox{yellow!40}{202} & \colorbox{red!40}{229} & 178 & 185 \\
        & Size$\downarrow$ & 99.41 & \colorbox{yellow!40}{12.40} & 304.24 & \colorbox{red!40}{9.79} & 204.59 & \colorbox{yellow!40}{60.76} & 253.82 & \colorbox{red!40}{27.36} & 87.97 & \colorbox{yellow!40}{12.80} & 554.02 & \colorbox{red!40}{5.95} & 52.74 & \colorbox{red!40}{5.44} & 181.72 & \colorbox{yellow!40}{6.09} \\
        \hline
        \multirow{6}{*}{L1}
        & PSNR$\uparrow$ & 20.90 & 20.85 & \colorbox{yellow!40}{27.56} & \colorbox{red!40}{27.64} & 18.77 & 20.98 & \colorbox{yellow!40}{26.17} & \colorbox{red!40}{27.13} & 24.28 & \colorbox{yellow!40}{27.38} & 21.00 & \colorbox{red!40}{28.44} & 17.30 & 15.57 & \colorbox{yellow!40}{24.00} & \colorbox{red!40}{24.25} \\
        & SSIM$\uparrow$ & 0.588 & 0.585 & \colorbox{yellow!40}{0.837} & \colorbox{red!40}{0.842} & 0.596 & 0.694 & \colorbox{yellow!40}{0.879} & \colorbox{red!40}{0.901} & 0.764 & \colorbox{yellow!40}{0.858} & 0.637 & \colorbox{red!40}{0.869} & 0.626 & 0.506 & \colorbox{yellow!40}{0.871} & \colorbox{red!40}{0.873} \\
        & LPIPS$\downarrow$ & 0.319 & 0.340 & \colorbox{red!40}{0.157} & \colorbox{yellow!40}{0.165} & 0.311 & 0.231 & \colorbox{yellow!40}{0.130} & \colorbox{red!40}{0.116} & 0.218 & \colorbox{yellow!40}{0.140} & 0.349 & \colorbox{red!40}{0.138} & 0.263 & 0.413 & \colorbox{red!40}{0.118} & \colorbox{yellow!40}{0.121} \\
        & FPS$\uparrow$ & \colorbox{red!40}{163} & 138 & 109 & \colorbox{yellow!40}{158} & \colorbox{red!40}{140} & 105 & 129 & \colorbox{yellow!40}{133} & \colorbox{red!40}{166} & 137 & 82 & \colorbox{yellow!40}{165} & \colorbox{red!40}{177} & 158 & 131 & \colorbox{yellow!40}{160} \\
        & Size$\downarrow$ & 198.82 & \colorbox{yellow!40}{45.77} & 730.14 & \colorbox{red!40}{27.65} & 409.18 & \colorbox{yellow!40}{197.18} & 470.34 & \colorbox{red!40}{59.04} & 175.94 & \colorbox{yellow!40}{48.72} & 1214.11 & \colorbox{red!40}{13.10} & 105.48 & \colorbox{yellow!40}{17.28} & 510.37 & \colorbox{red!40}{15.18} \\
        & $\Delta$Size$\downarrow$ & 99.41 & \colorbox{yellow!40}{33.37} & 425.90 & \colorbox{red!40}{17.86} & 204.59 & \colorbox{yellow!40}{136.42} & 216.52 & \colorbox{red!40}{31.68} & 87.97 & \colorbox{yellow!40}{35.92} & 660.09 & \colorbox{red!40}{7.15} & 52.74 & \colorbox{yellow!40}{11.84} & 328.65 & \colorbox{red!40}{9.09} \\
        \hline
        \multirow{6}{*}{L2}
        & PSNR$\uparrow$ & 25.18 & 25.34 & \colorbox{yellow!40}{26.84} & \colorbox{red!40}{26.95} & 23.87 & 24.53 & \colorbox{yellow!40}{25.12} & \colorbox{red!40}{26.84} & 28.29 & \colorbox{yellow!40}{29.57} & 22.55 & \colorbox{red!40}{29.76} & 20.69 & 21.34 & \colorbox{yellow!40}{23.91} & \colorbox{red!40}{24.11} \\
        & SSIM$\uparrow$ & 0.772 & 0.766 & \colorbox{yellow!40}{0.790} & \colorbox{red!40}{0.796} & 0.828 & \colorbox{yellow!40}{0.841} & 0.835 & \colorbox{red!40}{0.884} & 0.874 & \colorbox{red!40}{0.897} & 0.746 & \colorbox{yellow!40}{0.894} & \colorbox{yellow!40}{0.791} & 0.768 & \colorbox{red!40}{0.860} & \colorbox{red!40}{0.860} \\
        & LPIPS$\downarrow$ & 0.221 & 0.230 & \colorbox{red!40}{0.215} & \colorbox{yellow!40}{0.219} & 0.166 & \colorbox{yellow!40}{0.154} & 0.185 & \colorbox{red!40}{0.138} & 0.148 & \colorbox{red!40}{0.134} & 0.309 & \colorbox{yellow!40}{0.136} & 0.174 & 0.225 & \colorbox{yellow!40}{0.133} & \colorbox{red!40}{0.132} \\
        & FPS$\uparrow$ & \colorbox{red!40}{142} & 128 & 82 & \colorbox{yellow!40}{130} & \colorbox{yellow!40}{96} & \colorbox{yellow!40}{96} & 88 & \colorbox{red!40}{97} & \colorbox{yellow!40}{143} & 136 & 65 & \colorbox{red!40}{154} & \colorbox{red!40}{160} & 144 & 88 & \colorbox{yellow!40}{151} \\
        & Size$\downarrow$ & 397.63 & \colorbox{yellow!40}{87.72} & 1238.74 & \colorbox{red!40}{62.79} & 818.36 & \colorbox{yellow!40}{269.57} & 840.79 & \colorbox{red!40}{115.79} & 351.89 & \colorbox{yellow!40}{54.36} & 2067.72 & \colorbox{red!40}{22.91} & 210.95 & \colorbox{yellow!40}{33.20} & 1087.07 & \colorbox{red!40}{26.63} \\
        & $\Delta$Size$\downarrow$ & 198.81 & \colorbox{yellow!40}{41.95} & 508.60 & \colorbox{red!40}{35.14} & 409.18 & \colorbox{yellow!40}{72.39} & 370.45 & \colorbox{red!40}{56.75} & 175.95 & \colorbox{yellow!40}{5.64} & 853.61 & \colorbox{red!40}{9.81} & 105.47 & \colorbox{yellow!40}{15.92} & 576.70 & \colorbox{red!40}{11.45} \\
        \hline
        \multirow{6}{*}{L3}
        & PSNR$\uparrow$ & 26.86 & \colorbox{yellow!40}{27.10} & 26.19 & \colorbox{red!40}{27.17} & 26.95 & \colorbox{yellow!40}{27.34} & 23.95 & \colorbox{red!40}{27.61} & 29.37 & \colorbox{red!40}{30.01} & 24.22 & \colorbox{yellow!40}{29.99} & 23.22 & 22.80 & \colorbox{yellow!40}{23.27} & \colorbox{red!40}{23.47} \\
        & SSIM$\uparrow$ & \colorbox{red!40}{0.793} & \colorbox{yellow!40}{0.791} & 0.748 & 0.790 & 0.883 & \colorbox{yellow!40}{0.890} & 0.766 & \colorbox{red!40}{0.893} & \colorbox{yellow!40}{0.901} & 0.900 & 0.834 & \colorbox{red!40}{0.903} & 0.827 & 0.783 & \colorbox{yellow!40}{0.834} & \colorbox{red!40}{0.836} \\
        & LPIPS$\downarrow$ & \colorbox{yellow!40}{0.226} & 0.230 & 0.278 & \colorbox{red!40}{0.224} & 0.138 & \colorbox{yellow!40}{0.130} & 0.265 & \colorbox{red!40}{0.128} & \colorbox{yellow!40}{0.192} & 0.196 & 0.288 & \colorbox{red!40}{0.188} & \colorbox{red!40}{0.182} & 0.254 & 0.185 & \colorbox{yellow!40}{0.184} \\
        & FPS$\uparrow$ & 107 & \colorbox{yellow!40}{114} & 70 & \colorbox{red!40}{118} & 78 & \colorbox{yellow!40}{86} & 80 & \colorbox{red!40}{90} & 112 & \colorbox{yellow!40}{135} & 58 & \colorbox{red!40}{142} & 134 & \colorbox{red!40}{142} & 64 & \colorbox{yellow!40}{135} \\
        & Size$\downarrow$ & 795.26 & \colorbox{yellow!40}{103.61} & 1740.10 & \colorbox{red!40}{90.11} & 1636.72 & \colorbox{yellow!40}{313.61} & 1400.15 & \colorbox{red!40}{135.42} & 703.77 & \colorbox{yellow!40}{56.34} & 3032.85 & \colorbox{red!40}{37.65} & 421.91 & \colorbox{red!40}{37.82} & 2115.54 & \colorbox{yellow!40}{42.45} \\
        & $\Delta$Size$\downarrow$ & 397.63 & \colorbox{yellow!40}{15.89} & 501.36 & \colorbox{red!40}{27.32} & 818.36 & \colorbox{yellow!40}{44.04} & 559.36 & \colorbox{red!40}{19.63} & 351.88 & \colorbox{red!40}{1.98} & 965.13 & \colorbox{yellow!40}{14.74} & 210.96 & \colorbox{red!40}{4.62} & 1028.47 & \colorbox{yellow!40}{15.82} \\
        \hline
    \end{tabular}
\end{table*}

\subsection{Results and Evaluation}\label{sec:results}
In this section, we evaluate the core capability of our Iterative Gaussian Synopsis framework: its ability to create a compact, high-fidelity, multi-level LOD representation. We conduct a detailed comparison against prominent LOD-aware methods, including Octree-GS~\cite{ren2024octree} and LapisGS~\cite{Shi2024LapisGS}, as well as the adaptive 3DGS~\cite{kerbl3Dgaussians} approach, across four distinct Levels-of-Detail (L0 being the coarsest and L3 the finest). The evaluation covers both quantitative metrics and qualitative visual results to provide a comprehensive analysis.

\subsubsection*{\bf Quantitative Comparison}
The quantitative results across the Mip-NeRF360~\cite{barron2021mip}, BungeeNeRF~\cite{xiangli2022bungeenerf}, DeepBlending~\cite{hedman2018deep}, and Tanks\&Temples~\cite{knapitsch2017tanks} datasets are consolidated in \tablename~\ref{tab:lod_comparisons}. Because every method and every level is scored at its own proportionally scaled target resolution, the entries of \tablename~\ref{tab:lod_comparisons} are mutually consistent. The differing L0 to L3 trends across methods instead reflect level construction, namely purpose-built per-level models against progressively revealing one full-resolution model.

All results published in the original LapisGS paper are evaluated on training views without a train/test split, as clarified in the official repository discussion.\footnote{\raggedright LapisGS repository discussion, issue \#11: \url{https://github.com/nus-vv-streams/lapis-gs/issues/11}.} In contrast, all entries in \tablename~\ref{tab:lod_comparisons} use held-out test views and our own reproductions of the baselines. This difference in evaluation split explains the discrepancy with the published LapisGS values. For example, on DeepBlending, our reproduction reaches L0 SSIM 0.967 on training views against the published 0.966, while the same models score 0.573 on held-out views. The shortfall is concentrated in Dr.\ Johnson, whose base layer overfits its low-resolution training views, reaching 32.18\,dB on them but only 14.61\,dB on held-out views at L0, with artifacts that persist through the enhancement layers and leave an 11.1\,dB gap at L3. Playroom trains stably under the same configuration, so this is a sensitivity of the cascaded scheme rather than a uniform defect, and we compare against our own reproduction throughout.

Our method demonstrates superior performance in balancing rendering fidelity, model compactness, and operational efficiency across all evaluated levels. At the initial, coarser levels (L0 and L1), where model compactness is most critical for fast loading, our method outperforms the baselines in the size-to-quality trade-off, highlighting the efficiency of our unfolding scheme in creating a high-quality condensed representation. Furthermore, the delta size metrics validate the efficiency of the top-down unfolding scheme, in which upgrading to higher LODs requires transmitting only marginal data updates, such as compact MLP weights and codebook modulations, rather than redundant Gaussian parameters. The anchor-based backbone accounts for much of the storage reduction relative to raw 3DGS and LapisGS, as \tablename~\ref{tab:scaffold_ablation} shows against independently trained Scaffold-GS models. The unfolding scheme and the adaptation modules add the progressivity: every level stays close to an independently trained model in quality, and upgrading between adjacent levels costs only the $\Delta$Size payload.

The integrated FPS metrics also confirm that our framework maintains real-time rendering speeds throughout the progressive refinement process. While bottom-up baseline methods often suffer from performance degradation or structural redundancy at finer levels, our approach provides consistent rendering efficiency across the entire LOD hierarchy. These findings demonstrate that our top-down approach effectively preserves global fidelity, guarantees high-speed rendering, and enables level-specific adaptation with minimal transmission overhead.

\subsubsection*{\bf Qualitative Comparison}
\begin{figure*}
    \centering
    \includegraphics[width=\linewidth]{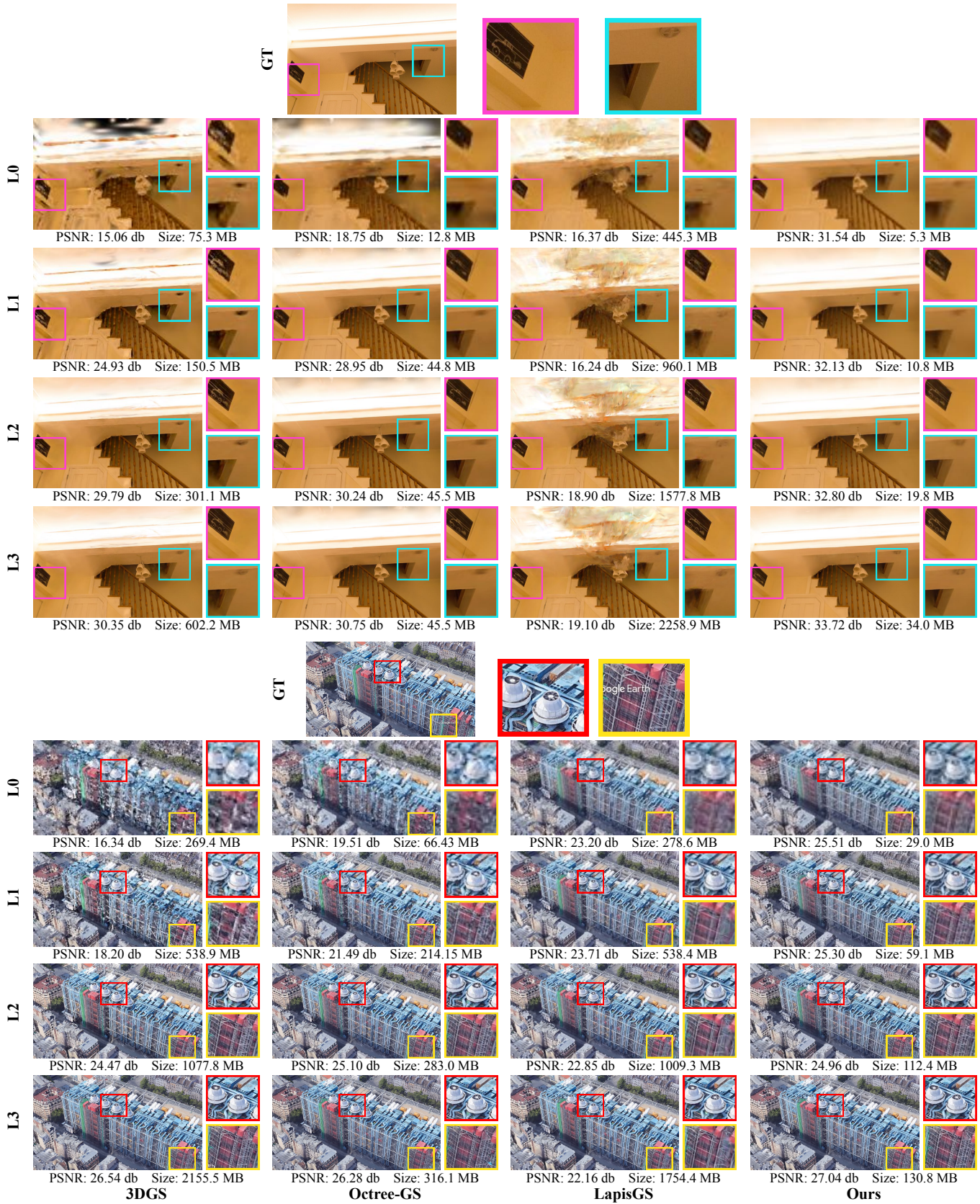}
    \caption{\textbf{Qualitative comparison of LOD rendering across different methods.} Our method yields superior visual quality (higher PSNR, and clearer details as shown in zoomed-in views) at each level, while requiring substantially less storage compared to all baseline methods.}
    \label{fig:comparisons}
\end{figure*}

The qualitative results presented in \figurename~\ref{fig:comparisons} visually corroborate our quantitative findings and highlight the practical advantages of our Iterative Gaussian Synopsis framework. At the coarsest levels (L0 and L1), our method provides a remarkably clear and structurally complete base representation while other methods exhibit distinct visual limitations. As the LOD increases, our method's ability to preserve fine details remains superior. The visual progression from L0 to L3 in our method is smooth and coherent, with each level adding meaningful detail. The same base-layer overfitting is visible in \figurename~\ref{fig:comparisons} as persistent floaters in the LapisGS renderings on DeepBlending. Our top-down unfolding scheme circumvents this issue by ensuring each LOD is a principled simplification of a high-quality parent model.

\begin{figure*}[!t]
    \centering
    \subfloat{\includegraphics[width=0.32\textwidth]{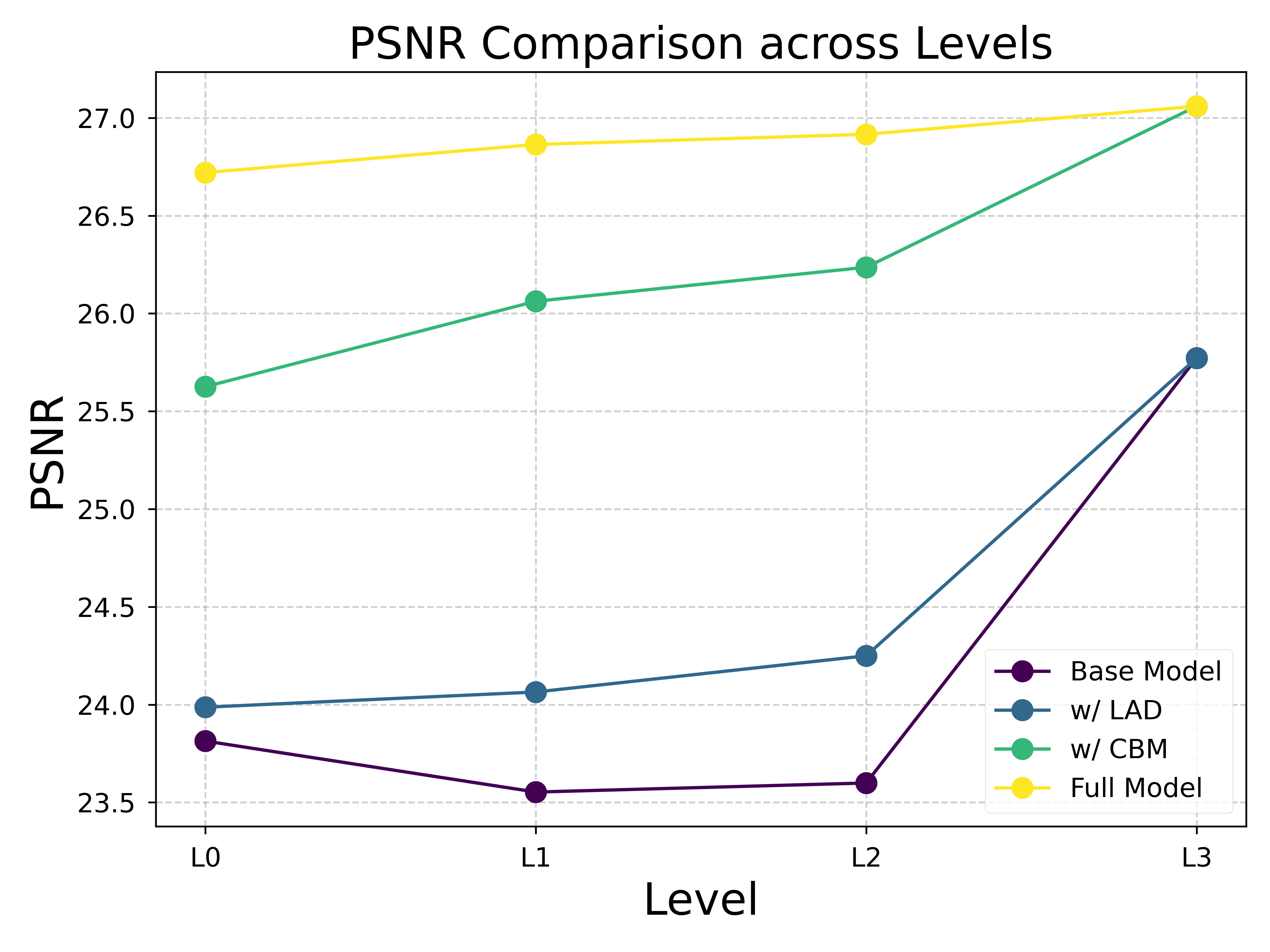}}
    \hfil
    \subfloat{\includegraphics[width=0.32\textwidth]{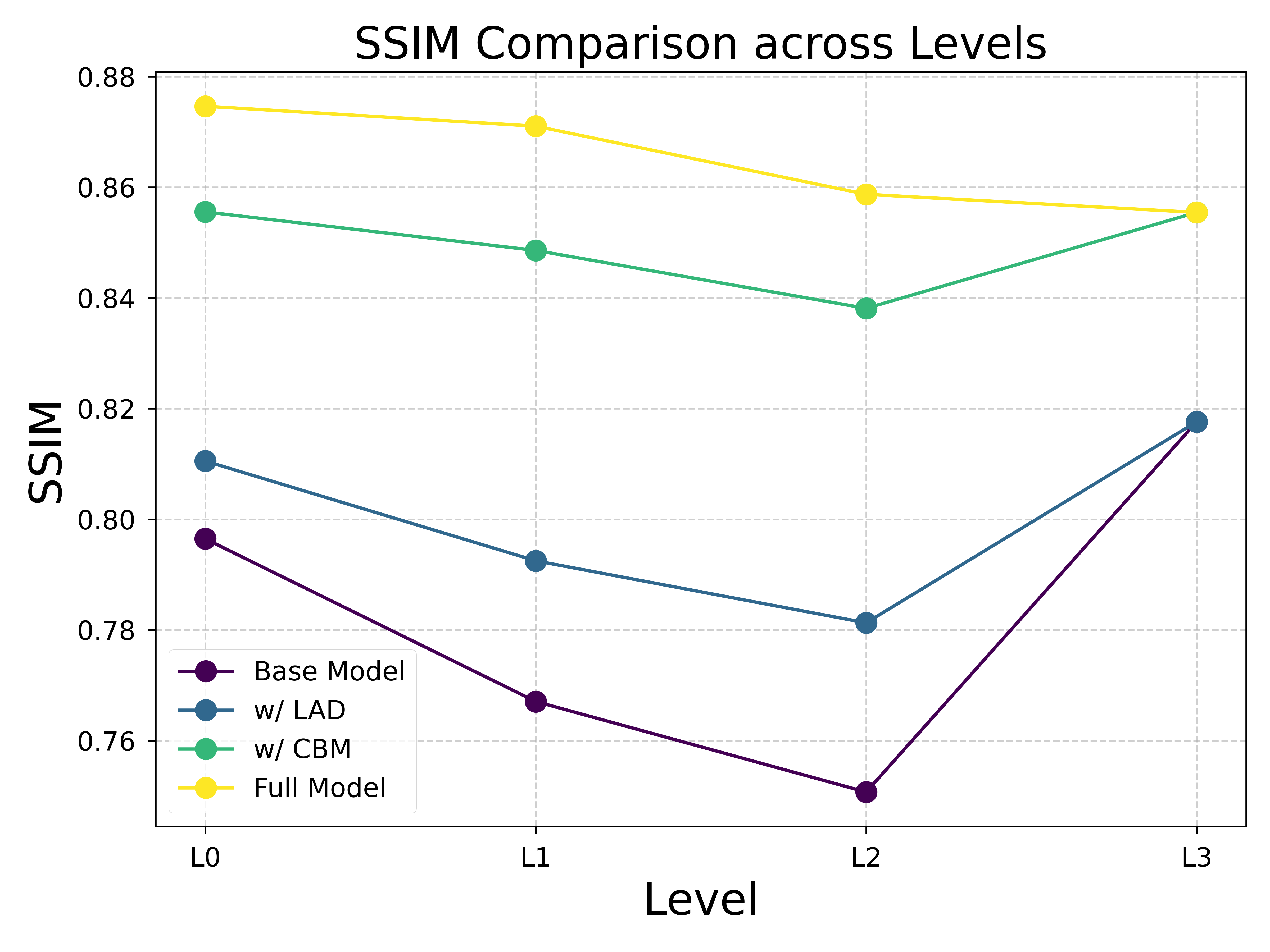}}
    \hfil
    \subfloat{\includegraphics[width=0.32\textwidth]{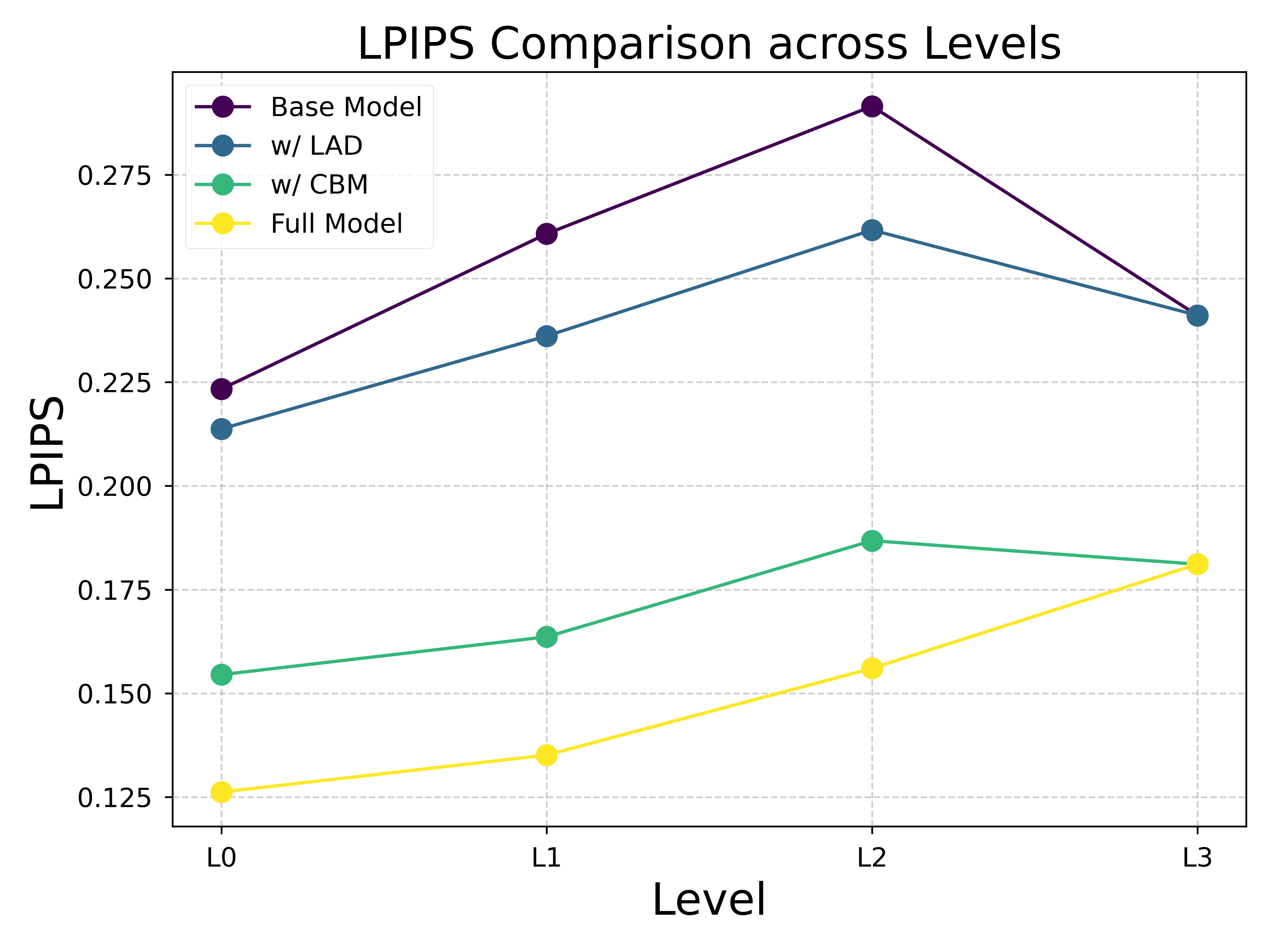}}
    \caption{\textbf{Quantitative results of the ablation study for our adaptation modules.} Across all metrics (PSNR, SSIM, LPIPS) and at every level, the full integration of both modules yields the highest fidelity, validating the effectiveness of our framework.}
    \label{fig:ablation_charts}
\end{figure*}

\begin{table}[h]
\centering
\caption{Architectural Analysis across multiple datasets. Our unified, progressive framework is compared against Scaffold-GS models trained independently at corresponding scaled resolutions.}
\label{tab:scaffold_ablation}
\scriptsize
\renewcommand{\arraystretch}{1.2}
\setlength\tabcolsep{1.8pt}
\begin{tabular}{l|c|cccc|cccc}
\hline
\multirow{2}{*}{} & \multirow{2}{*}{\textbf{Level}} & \multicolumn{4}{c|}{\textbf{Scaffold-GS (Independent)}} & \multicolumn{4}{c}{\textbf{Ours (Unified Progressive)}} \\ \cline{3-10}
 &  & PSNR & SSIM  & Tr.(min) & Cum. Size & PSNR & SSIM  & Tr.(min) & Cum. Size \\ \hline
\multirow{4}{*}{\textbf{Mip.}}
 & L0  & 28.75 & 0.874 & 12.2 & 89.80 MB & 28.02 & 0.860 & 4.5 & 9.79 MB \\
 & L1 & 28.15 & 0.850 & 17.3 & 223.23 MB & 27.64 & 0.842 & 9.0 & 27.65 MB \\
 & L2 & 27.23 & 0.813 & 23.1 & 388.47 MB & 26.95 & 0.796 & 12.4 & 62.79 MB \\
 & L3 & 27.09 & 0.793 & 39.0 & 560.27 MB & 27.17 & 0.790 & 32.2 & 90.11 MB \\ \hline
\multirow{4}{*}{\textbf{DB}}
 & L0 & 29.13 & 0.885 & 7.4 & 39.10 MB & 27.95 & 0.861 & 5.0 & 5.95 MB \\
 & L1 & 28.75 & 0.898 & 9.0 & 86.31 MB & 28.44 & 0.869 & 5.1 & 13.10 MB \\
 & L2 & 30.06 & 0.899 & 16.7 & 139.94 MB & 29.76 & 0.894 & 9.4 & 22.91 MB \\
 & L3 & 29.97 & 0.901 & 28.2 & 193.79 MB & 29.99 & 0.903 & 19.1 & 37.65 MB \\ \hline
\multirow{4}{*}{\textbf{TnT}}
 & L0 & 25.11 & 0.881 & 10.5 & 42.10 MB & 23.68 & 0.865 & 3.9 & 6.09 MB \\
 & L1 & 24.48 & 0.876 & 18.0 & 91.45 MB & 24.25 & 0.873 & 7.1 & 15.18 MB \\
 & L2 & 24.27 & 0.869 & 17.1 & 151.40 MB & 24.11 & 0.860 & 8.0 & 26.63 MB \\
 & L3 & 22.92 & 0.841 & 21.6 & 225.47 MB & 23.47 & 0.836 & 14.2 & 42.45 MB \\ \hline
\end{tabular}\end{table}

\subsection{Architectural Analysis}
Beyond our level-specific adaptation modules, our foundational architecture diverges from the standard Scaffold-GS~\cite{scaffoldgs} design by utilizing multi-scale grids and a shared Anchor Codebook rather than assigning independent learnable features to each anchor. To validate this design choice, we evaluated an independent-feature baseline by training separate Scaffold-GS models at the scaled resolutions corresponding to our LODs.

\tablename~\ref{tab:scaffold_ablation} compares our unified top-down representation against these independently trained Scaffold-GS models. Because the Scaffold-GS models are unconstrained by a multi-level hierarchy and are optimized entirely from scratch for each specific resolution, they establish a theoretical upper bound for static fidelity. As shown, our method achieves highly competitive visual quality, trailing the independent baseline by only a marginal gap in lower LODs.

In a progressive streaming context, however, independent models multiply both the training compute and the payload a client must download and decode, whereas our shared codebook and grids extract all LODs from a single optimization and upgrade between levels with only lightweight codebook and MLP updates, as \tablename~\ref{tab:scaffold_ablation} quantifies.

\subsection{Ablation Studies}
To validate the effectiveness and necessity of the core components of our Iterative Gaussian Synopsis framework, we conducted a series of ablation studies. We specifically analyze the performance impact of our two key level-specific adaptation modules: Level-aware Decoding (LAD) and  Coherent Basis Modulation (CBM). We compare our full method against variants where these modules are incrementally added to a foundational baseline. The configurations evaluated are:

\begin{itemize}[leftmargin=*]
    \item {Base Model:} Our foundational framework without the level-specific adaptations, relying only on the pruned anchor structure for each LOD.
    \item {w/ LAD:} The Base Model augmented with \textit{only} the LAD module. To strictly decouple this from CBM, the local feature branch is completely discarded. The level-specific MLPs rely exclusively on the downsampled hierarchical grid features to predict Gaussian attributes. This configuration strictly isolates the decoding capability of LAD from any local feature assistance, demonstrating its effectiveness in interpreting multi-scale global context independently.
    \item {w/ CBM:} The Base Model augmented with \textit{only} the Coherent Basis Modulation module. In this configuration, the level-adapted local features retrieved from the modulated codebook are fused with the downsampled hierarchical features, but they are decoded using the \textit{frozen, foundational MLPs} established during the initial full-fidelity training. This demonstrates the performance gain achieved solely by adapting the local feature bases for a specific LOD, without specializing the decoding functions.
    \item {Full Model:} Our complete proposed method, incorporating \textit{both} CBM and LAD on top of the base framework.
\end{itemize}

The quantitative results are visualized in \figurename~\ref{fig:ablation_charts}. The charts clearly show that the ``Base Model'' establishes a performance lower bound. From there, adding either of our proposed adaptation modules individually provides a significant improvement, confirming their individual contributions to enhancing rendering quality. Crucially, the ``Full Model'' consistently achieves the best performance across all metrics (PSNR, SSIM, LPIPS) and at every LOD, indicating a strong synergistic effect. This demonstrates that while both adapting the anchor features (via CBM) and specializing their interpretation (via LAD) are beneficial, they work best in tandem to produce the highest-fidelity results.

\section{Conclusion}
In this paper, we introduced \textit{Iterative Gaussian Synopsis}, a novel framework that addresses key challenges in deploying large-scale 3DGS models for progressive and adaptive rendering. Our approach implements a top-down ``unfolding'' scheme that constructs a multi-level LOD hierarchy through a learnable, group-aware anchor pruning strategy. By combining a structurally coherent representation with lightweight, level-specific adaptation modules, our method maximizes inter-layer reusability and enables significant detail refinement with minimal data overhead. Experimental results demonstrate that our framework achieves state-of-the-art performance in balancing compactness and high-fidelity rendering across all LODs, offering a practical and scalable solution for real-time 3DGS deployment.

\textbf{Limitations and Future Work:} Despite its strengths, the framework introduces additional training complexity and, like most current methods, is limited to static scenes. Furthermore, while our shared anchor representation naturally minimizes spatial discontinuity during LOD transitions compared to independent-parameter baselines, the discrete nature of the hierarchy can still introduce mild visual ``popping'' artifacts as new anchors activate. These constraints suggest promising directions for future work. Potential extensions include exploring continuous transition mechanisms or client-side interpolation techniques to fully eliminate popping, adapting the unfolding mechanism to dynamic content, incorporating perceptually guided pruning to better align LOD transitions with human visual sensitivity, and integrating the framework into an end-to-end adaptive streaming pipeline for evaluation under real-world network conditions.

\bibliographystyle{IEEEtran}
\bibliography{IEEEabrv, mybib}

\end{document}